\documentclass[10pt,twocolumn,letterpaper]{article}

\usepackage{cvpr}              

\usepackage{graphicx}
\usepackage{amsmath}
\usepackage{amssymb}
\usepackage{booktabs}
\usepackage{times}
\usepackage{epsfig}
\usepackage{graphicx}

\newcommand{\bs}{\boldsymbol}
\usepackage[pagebackref,breaklinks,colorlinks]{hyperref}

\usepackage[capitalize]{cleveref}
\crefname{section}{Sec.}{Secs.}
\Crefname{section}{Section}{Sections}
\Crefname{table}{Table}{Tables}
\crefname{table}{Tab.}{Tabs.}

\def\cvprPaperID{2686} 
\def\confName{CVPR}
\def\confYear{2022}

\begin{document}

\title{3D Shape Reconstruction from 2D Images with Disentangled Attribute Flow}

\author{Xin Wen\textsuperscript{1,2,}\thanks{Equal contribution.}, Junsheng Zhou$^{1,*}$, Yu-Shen Liu$^1$\thanks{The corresponding author is Yu-Shen Liu. This work was supported by National Key R\&D Program of China (2018YFB0505400, 2020YFF0304100), the National Natural Science Foundation of China (62072268), and in part by Tsinghua-Kuaishou Institute of Future Media Data.}, Hua Su$^3$, Zhen Dong$^4$, Zhizhong Han$^5$\\
School of Software, BNRist, Tsinghua University, Beijing, China$^1$\\
JD Logistics, JD.com, Beijing, China$^2$\hspace {5 mm}Kuaishou Technology, Beijing, China$^3$\\
Wuhan University, Wuhan, China$^4$\\
Department of Computer Science, Wayne State University, Detroit, USA$^5$\\
{\tt\small wenxin16@jd.com\hspace {3mm}zhoujs21@mails.tsinghua.edu.cn\hspace {3mm}liuyushen@tsinghua.edu.cn}\\
{\tt\small shlw@kuaishou.com\hspace {3mm}dongzhenwhu@whu.edu.cn\hspace {3mm}h312h@wayne.edu}
}

\maketitle

\begin{abstract}
Reconstructing 3D shape from a single 2D image is a challenging task, which needs to estimate the detailed 3D structures based on the semantic attributes from 2D image.
So far, most of the previous methods still struggle to extract semantic attributes for 3D reconstruction task. Since the semantic attributes of a single image are usually implicit and entangled with each other, it is still challenging to reconstruct 3D shape with detailed semantic structures represented by the input image.
To address this problem, we propose 3DAttriFlow to disentangle and extract semantic attributes through different semantic levels in the input images.
These disentangled semantic attributes will be integrated into the 3D shape reconstruction process, which can provide definite guidance to the reconstruction of specific attribute on 3D shape. As a result, the 3D decoder can explicitly capture high-level semantic features at the bottom of the network, and utilize low-level features at the top of the network, which allows to reconstruct more accurate 3D shapes.
Note that the explicit disentangling is learned without extra labels, where the only supervision used in our training is the input image and its corresponding 3D shape. Our comprehensive experiments on ShapeNet dataset demonstrate that 3DAttriFlow outperforms the state-of-the-art shape reconstruction methods, and we also validate its generalization ability on shape completion task. Code is available at \url{https://github.com/junshengzhou/3DAttriFlow}.

\end{abstract}

\section{Introduction}
\label{sec:introduction}

Reconstructing a 3D shape from a 2D image (2D-to-3D reconstruction) is a crucial task for bridging the gap between the 2D and 3D visual understanding. The typical paradigm is to firstly capture the semantic features of the 2D images through an image encoder, and then correctly reconstruct them in 3D space through a 3D decoder. Among the multiple representation forms of 3D shapes (i.e. voxel, point cloud and mesh), this paper mainly focuses on reconstructing 3D point cloud from the input image, due to its lightweight storage consumption and capability of representing various complicated shapes.

\begin{figure*}[!t]
  \centering
  \includegraphics[width=\textwidth]{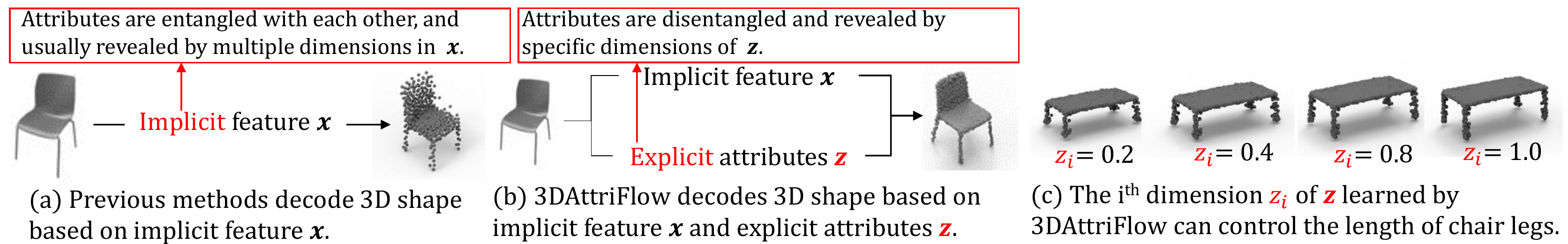}\vspace{-0.2cm}
  \caption{Comparison between previous methods (a) and our 3DAttriFlow (b). Besides the implicit feature $\bs{x}$ of input image, 3DAttriFlow learns an additional attribute code $\bs{z}$, which can reveal some hints about more specific semantic attributes of 3D shape in (c).
  }
  \label{fig:problem}
\end{figure*}

As addressed by the typical paradigm of most previous methods \cite{xie2019pix2vox,wang2018pixel2mesh,wen2019pixel2mesh++,scheder2019occupancy}, the key of 2D-to-3D reconstruction is how to precisely interpret the semantic attribute from images into the 3D space. Thanks to the recent progress of 2D computer vision, there are many well-known methods (e.g. AlexNet\cite{krizhevsky2012imagenet}, VGG\cite{vgg2015} and ResNet\cite{he2016deep}) to encode semantic attributes into image features, and their efficiencies have also been proved by a wide range of cross-modal tasks (e.g. image captioning \cite{xu2015show,vinyals2015show}, cross-modal retrieval \cite{wang2017adversarial,zhen2019deep}). However, for the research of 2D-to-3D reconstruction, how to interpret visual information from 2D domain to 3D domain for accurate 3D reconstruction still remains a difficult task.
Because most previous methods \cite{xie2019pix2vox,xie2020pix2vox++,wang2018adaptive,wang2018pixel2mesh,wen2019pixel2mesh++,scheder2019occupancy} only rely on the feature channels (e.g. element-wise add, feature concatenation and attention mechanism) to convey the visual information from the image encoder to the 3D decoder, which only contains implicit geometric information with limited semantic attributes as the guidance to shape reconstruction.
For example, an overall geometric information such as the number of legs will determine the table to have three or four legs. Such geometric information can be easily noticed and reconstructed by the decoder. 
On the other hand, the detailed semantic attributes like the length or bending of legs will specifically determine the detailed shape of these legs. However, since these semantic attributes are deeply entangled with each other in image features, they can hardly be noticed by the decoder during the reconstruction process. 

Moreover, semantic attributes are usually distributed at various semantic levels, and entangle with each other throughout the pyramidal hierarchy of image encoder. As a result, they can hardly be fully exploited through implicit feature channels. 
As a result, the previous methods usually suffer from guiding the decoder to reconstruct various visual information extracted by the encoder, which leads to the insufficient usage of semantic features for predicting 3D shapes in the previous methods.

A straightforward solution to this problem is to build numbers of feature channels between the decoder and all the network layers in the encoder, which will increase the cost of tremendous computational time and network complexity. 
On the other hand, as proved by many methods of image-to-image translation (e.g. image super-resolution \cite{xiao2020neural,wang2018esrgan}, image style transferring \cite{zhu2017unpaired}), we notice that the global feature is able to encode most of the semantic attributes for a single image, as they can be used for high quality image generation/restoration task. 
Therefore, a promising solution is to explore deeply into the global features extracted from the 2D images, and decode the abundant semantic attributes embedded in the global features, which may provide more detailed and definite guidance to the reconstruction process of 3D shapes.
Following the above-mentioned intuition, we propose a novel neural network, named 3DAttriFlow, to decompose the semantic attributes from the 2D image, and utilize these semantic attributes for 3D shape reconstruction in a controllable way.

Specifically, as shown in Figure \ref{fig:problem}, previous methods (Figure \ref{fig:problem}(a)) usually learn to reconstruct 3D shapes from an implicit image feature. In contrast, 3DAttriFlow tries to decompose an attribute code (Figure \ref{fig:problem}(b)) as hints to capture some specific semantic attributes (Figure \ref{fig:problem}(c)). 
Such process is accomplished by the \emph{attribute flow pipe} proposed in 3DAttriFlow.
By piping semantic attributes hints into the hierarchical generation process of point clouds through the attribute flow pipe, the decoder is able to selectively interpret semantic attributes following the hierarchy of semantic levels. 

Our idea is inspired by the recent generative method of EigenGAN \cite{he2021eigengan}, which learns to manipulate explicit semantic attributes of human faces in an unsupervised way. 
However, due to the discrete nature of point clouds, the coordinates of points are merely organized in an unordered manner, which is in contrast with the image pixels arranged in an ordered grid structure. Such nature of point clouds makes the location of each point unpredictable during the generation process, until the 3-dimensional coordinates are finally revealed at the end of the decoder. 
Therefore, a direct implementation of EigenGAN \cite{he2021eigengan} based decoder may result in failure, because the network cannot accurately predict the semantic attribute for a specific point without knowing its location. 
To address this problem, we propose the \emph{deformation pipe} as the solution, which follows the idea of PMP-Net\cite{wen2021pmp} to reconsider the shape generation process as a shape deformation process. That is, each point is first assigned a prior location in 3D space, and then moved to their destination to regroup as a new shape. Specifically, 3DAttriFlow moves the point cloud sampled from a 3D sphere into the target shape indicated by the 2D images.
In all, our main contributions are summarized as follows.
\begin{itemize}
    \item We propose a novel deep network, named 3DAttriFlow, for reconstructing high-quality 3D shapes from single 2D images. Compared with the previous methods, 3DAttriFlow can interpret explicit semantic attributes from images, and effectively use them to guide the decoder for detailed and high-quality 2D-to-3D shape reconstruction.
    \item We propose the attribute flow pipe to explicitly disentangle the semantic attributes embedded in the global feature of 2D image, which can provide definite guidance about the detailed reconstruction of semantic attributes to the 3D decoder, leading to more accurate prediction of 3D shape in terms of both overall and detailed shape structures.
    \item We propose the deformation pipe to offer the location priors to attribute flow pipe, where the extracted semantic attributes can be assigned to a specific point by leveraging the location of that point. As a result, 3DAttriFlow avoids the problem of assigning semantics to unordered data, and allows more accurate feature integration between the attribute flow pipe and the deformation pipe.
\end{itemize}

\section{Related Work}
\label{sec:related_work}
Recent improvement of 3D representation learning \cite{han20193dviewgraph,han2019parts,wen2020point2spatialcapsule,han2019seqviews2seqlabels,liu2020lrc,Wen2020MM}, reconstruction \cite{ma2022surface,ma2022recon,li2022learning,Han2020ECCV,Han2020Sdrwr,Han2020TIP,wen2022pmp} and completion \cite{wen2020sa,xin2021c4c} in 3D computer vision field have boosted the research of reconstructing 3D shapes from 2D images, which can be categorized according to the number of input images as: single-view 3D shape reconstruction\cite{wang2018adaptive,wang2018pixel2mesh,wen2019pixel2mesh++,scheder2019occupancy,groueix2018papier} and multi-view 3D shape reconstruction\cite{xie2019pix2vox,wen2019pixel2mesh++,xie2020pix2vox++}. On the other hand, according to different representation forms of 3D shapes, the related work can also be categorized as voxel-based 3D shape reconstruction\cite{xie2019pix2vox,xie2020pix2vox++,scheder2019occupancy,wang2018adaptive}, point cloud based 3D shape reconstruction\cite{groueix2018papier,fan2017point,wang2019mvpnet,lin2018learning} and mesh-based 3D shape reconstruction\cite{wang2018pixel2mesh,wen2019pixel2mesh++}. Specifically, the proposed 3DAttriFlow in this paper belongs to the single-view 3D shape reconstruction, which is based on point clouds. The discussion of related work will be organized according to the output forms of 3D shape for convenience.

\textbf{Point cloud based methods.} With the rapid development of point cloud representation learning\cite{qi2017pointnet,qi2017pointnet2,li2018pointcnn,wang2019dynamic,liu2020lrc}, which is triggered by the pioneering work of PointNet\cite{qi2017pointnet}, point cloud generation has been widely studied in recent years, and boosted the research of reconstructing point clouds from 2D images.
Most of the point cloud based methods \cite{groueix2018papier,fan2017point,jiang2018gal,navaneet2019capnet,2DProjectionMatching} follow the generative way to predict the point coordinates based on the 2D images, where their efforts are made either to improve the feature communications between image encoder and 3D shape decoder \cite{fan2017point}, or impose extra supervision/constraint on the generated point clouds \cite{jiang2018gal,navaneet2019capnet,zhang2021view}. 

\textbf{Voxel/mesh based methods.} 
As for voxel-based reconstruction methods, the grid structure of 3D voxels is naturally applied in convolutional neural network, which simplifies the problem as translating 2D grid data to 3D grid data. 
Typical practice along this line is to directly utilize the CNN structure in both 2D and 3D domain, which aims to extract 2D grid feature from the input image, and reconstruct the corresponding 3D grid shape. Typical methods like 3DR2N2\cite{choy2016r2n2}, Pix2Vox\cite{xie2019pix2vox} and Pix2Vox++\cite{xie2020pix2vox++} have comprehensively explored the 3D reconstruction performance using single or multiple images as inputs.
However, suffering from the cubic growth of input voxel data, the resolution for voxel data is usually limited, while further increasing the resolution will lead to unacceptable computational cost.
As for mesh based methods, most of them follow the idea of deforming from a prior shape. For example, Pixel2Mesh\cite{wang2018pixel2mesh} and its successor Pixel2Mesh++\cite{wen2019pixel2mesh++} consider to deform an ellipsoid mesh into a target shape, which is combined with a multi-stage fusion strategy to introduce image features into the mesh deformation network. Li et al. \cite{li2020self} further extend such framework to capture the semantic part of object in 2D images. Pan et al.\cite{pan2019deep} improve the ability to generate complex shape by deforming mesh while modifying its typology.
However, the intersection of meshes and the hypothesis of manifold surface will hinder the generation of 3D shape with inner or irregular structures. 

\textbf{Discussion.} The reconstruction of 3D shapes from 2D images requires the deep understanding of semantic attributes in 2D images, and the correct interpretation of semantic attributes in 3D space. The above-mentioned methods either choose to directly decode the 3D shape from a global feature, or rely on feature channels to bridge the network layers between image encoder and the shape decoder. 
The problem is that, all these practices can only convey the implicit features from 2D images to 3D shapes, resulting in ambiguous guidance to reconstruct specific and detailed semantic attributes of 3D shape. 
Different from these previous methods, 3DAttriFlow proposes the solution to directly decompose the semantic attributes from the image feature, and integrate them into the shape reconstruction process, which can offer a definite guidance to the reconstruction of specific semantic attribute according to the 2D image. 
Moreover, the ability of attribute decomposition in 3DAttriFlow enables the decoder to flexibly reconstruct the semantic attributes following the hierarchy of semantic levels, which is in contrast to the network with fixed channels that only allows decoder to learn from fixed layers of encoder.

\begin{figure*}[!t]
  \centering
  \includegraphics[width=0.95\textwidth]{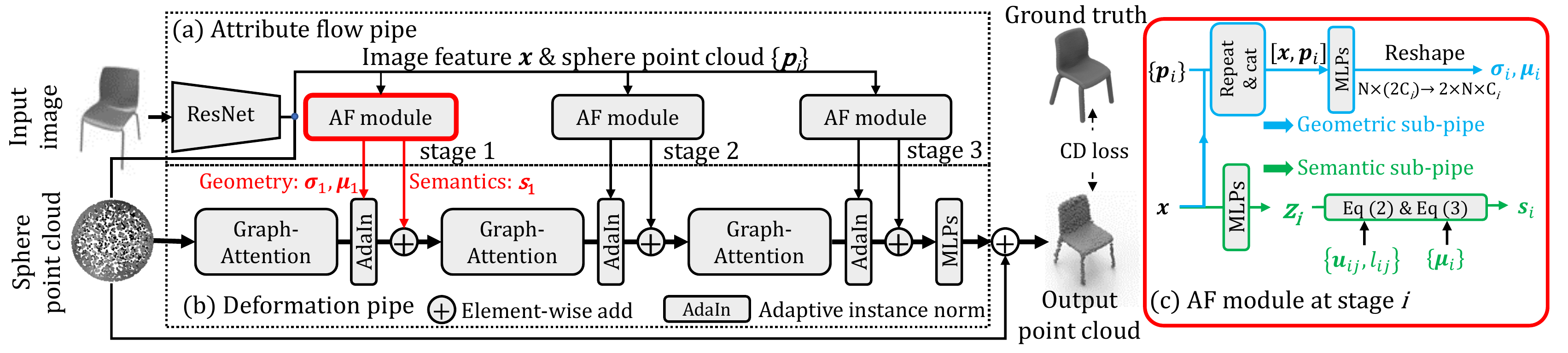}\vspace{-0.2cm}
  \caption{The overall architecture of 3DAttriFlow. 3DAttriFlow consists of two pipelines: (a) the attribute flow (AF) pipe extracts geometric code and semantic features based on the input image and initial sphere point cloud; (b) the deformation pipe deforms the point cloud according to the output of attribute flow pipe into the target shape. The detailed structure of AF module is shown in (c).
  }
  \label{fig:overview}
\end{figure*}
\section{Architecture of 3DAttriFlow}

The overall architecture of 3DAttriFlow is shown in Figure \ref{fig:overview}, which reconstructs a 3D point cloud with $N$ points according to the input image. 3DAttriFlow mainly consists of two pipelines as follows. 
(1) The \emph{attribute flow pipe} (see Figure \ref{fig:overview}(a)) serves to disentangle the semantic attributes from the input feature, which is usually the global feature extracted by an image encoder. 
(2) The \emph{deformation pipe} (see Figure \ref{fig:overview}(b)) serves to deform the initial point cloud sampled from a 3D sphere into the target shape, which is guided by the semantic attributes from the attribute flow pipe. The  structures for each pipeline are detailed below.
\subsection{Attribute Flow Pipe}
As shown in Figure \ref{fig:overview}(a), the attribute flow pipe aims to extract geometric code $\{\bs{\sigma}, \bs{\mu}\}$ and semantic features $\bs{s}_i$ step-by-step from the image feature $\bs{x}$ and sphere point cloud $\{\bs{p}_k\}$, where $i$ denotes the $\bs{i}$\textsuperscript{th} step. 
Then, the extracted feature and code will be integrated into the deformation pipe to guide the deformation of the spherical point cloud $\{\bs{p}_k\}$. 
The basic architecture of attribute flow pipe consists of a feature extractor and three attribute flow modules (AF module).
Specifically, for the input image, 3DAttriFlow uses ResNet18 to extract an image feature $\bs{x}$ from the input image. 
Then, the AF module extracts and interprets the visual information from the image feature $\bs{x}$ to the geometric information and semantic attributes, which is accomplished by the geometric sub-pipe and the semantic sub-pipe, as shown in Figure \ref{fig:overview}(c).

\textbf{Geometric sub-pipe.} 
The geometric sub-pipe aims to interpret the overall visual information from images into the geometric information, which can be utilized for 3D shape reconstruction by the deformation pipe.
Inspired by the style transferring based generative methods \cite{li2021sp,karras2019style}, which learns the \emph{local styles} from the latent random vector, we propose to interpret the visual information encoded by image feature $\bs{x}$ into the \emph{geometric styles} $\{\bs{\sigma}_i,\bs{\mu}_i\}$, which is according to the location prior given by the initial point cloud $\{\bs{p}_k\}$.
As shown by the geometric sub-pipe in Figure \ref{fig:overview}(c), at stage $i$, the image feature $\bs{x}$ is first repeated and concatenated with location priors $\{\bs{p}_k\}$ as $\{[\bs{x}:\bs{p}_k]\}$, where ``:'' denotes the feature concatenation. 
Then, followed by several multi-layer perceptrons (MLPs) and reshape operation, the image feature coupled with location priors is interpreted as geometric styles $\{\bs{\sigma}_i|\bs{\sigma}_i\in \mathbb{R}^{N\times C_i}\}$ and $\{\bs{\mu}_i|\bs{\mu}_i\in \mathbb{R}^{N\times C_i}\}$, where $C_i$ denotes the dimension of point features in deformation pipe at stage $i$.

\textbf{Semantic sub-pipe.} The semantic sub-pipe aims to decompose explicit semantic attributes from the image feature $\bs{x}$, and represent them by the activation at certain dimension of attribute code $\bs{z}$. 
As a result, the deformation pipe can produce a precise 3D semantic attribute under the definite guidance given by the attribute code. 
Specifically, as shown in the lower-branch of Figure \ref{fig:overview}(c), at stage $i$, the semantic sub-pipe first squeezes the image feature $\bs{x}$ into attribute code $\bs{z}_i$ as:
\begin{equation}
    \bs{z}_i = \phi(\bs{x}|\theta_i),
\end{equation}
where $\phi$ denotes the MLP layer, and $\theta_i$ denotes the weights of MLP layer for generating $\bs{z}_i$.
According to He et al.\cite{he2021eigengan}, for the activation $z_{ij}$ at $j$\textsuperscript{th} dimension of the attribute code $\bs{z}_i$, an orthogonal  basis $\bs{u}_{ij}\in \mathbb{R}^{N \times C_i}$ from a linear subspace $\mathcal{U}_i=\{\bs{u}_{ij}\}$ will be use to discover the semantic attribute $\bs{\hat{z}_{ij}}$ lying behind $z_{ij}$ as:
\begin{equation}
    \bs{\hat{z}_{ij}} = z_{ij}l_{ij}\bs{u}_{ij},
\end{equation}
where $l_{ij}$ is a learnable weight denoting the significance of the semantic attribute discovered by the orthogonal basis $\bs{u}_{ij}$. By adding the semantic attributes $\bs{\hat{z}_j}$ across all dimensions of attribute code $\bs{z}_i$, the semantic sub-pipe outputs the semantic feature $\bs{s}_i$ encoded with explicit attribute information, which is formulated as:
\begin{equation}
    \bs{s}_i = \sum_{j}\bs{\hat{z}}_{ij} + \bs{b}_i,
\end{equation}
where $\bs{b}_i$ is a learnable bias. The semantic feature $\bs{s}_i$ will be flowed into the deformation pipe to guide the reconstruction of 3D semantic attributes.

\subsection{Deformation Pipe}
The architecture of the deformation pipe is shown in Figure \ref{fig:overview}(b). The input at the bottom of the deformation pipe is a point set $\mathcal{P} = \{\bs{p}_i\}$, which is uniformly sampled from a 3D sphere. Note that we choose sphere as a starting shape because each point on the sphere can be regarded as a L2-regularized vector, which guarantees an isotropic shape prior input to the network.
The output at the top of the deformation pipe is a set of displacement vector $\{\Delta \bs{p}_i\}$. 
The output of the deformation pipe is a deformed point set $\mathcal{P}^o=\{(\bs{p}_i+\Delta \bs{p}_i)\}$, which has the same shape as target point cloud $\mathcal{P}^t = \{\bs{p}^t_j\}$.


\begin{table*}[!t]\small
\setlength{\abovecaptionskip}{-0.cm}
\centering
\caption{2D-to-3D reconstruction on ShapeNet dataset in terms of per-point L1 Chamfer distance $\times 10^2$ (lower is better).}
\resizebox{\textwidth}{!}{
\begin{tabular}{l|c|ccccccccccccc}
\toprule
Methods &Average  &Plane    &Bench  &Cabinet   &Car   &Chair   &Display    &Lamp    &Loud.   &Rifle &Sofa  &Table  &Tele. &Vessel   \\ \midrule
3DR2N2 \cite{choy2016r2n2}   &5.41   &4.94   &4.80  &4.25  &4.73    &5.75   &5.85    &10.64    &5.96    &4.02    &4.72  &5.29   &4.37   &5.07   \\
PSGN  \cite{fan2017point}   &4.07 &2.78  &3.73    &4.12    &3.27    &4.68    &4.74    &5.60    &5.62    &2.53 &4.44 &3.81 &3.81 &3.84   \\
Pixel2mesh  \cite{wang2018pixel2mesh}   &5.27  &5.36 &5.14    &4.85    &4.69    &5.77    &5.28    &6.87    &6.17    &4.21   &5.34   &5.13   &4.22   &5.48   \\
AtlasNet \cite{groueix2018papier}  &3.59 &2.60    &3.20    &3.66    &3.07    &4.09    &4.16    &4.98    &4.91  &2.20  &3.80   &3.36   &3.20   &3.40   \\
OccNet \cite{scheder2019occupancy}   &4.15   &3.19   &3.31    &3.54    &3.69    &4.08    &4.84    &7.55    &5.47    &2.97  &3.97   &3.74   &3.16   &4.43   \\
\midrule
3DAttriFlow(Ours) &\textbf{3.02}    &\textbf{2.11}  &\textbf{2.71}   &\textbf{2.66}   &\textbf{2.50}  &\textbf{3.33}  &\textbf{3.60}  &\textbf{4.55}  &\textbf{4.16}  &\textbf{1.94}  &\textbf{3.24}  &\textbf{2.85}  &\textbf{2.66}  &\textbf{2.96}   \\
\bottomrule
\end{tabular}
}
\label{table:2D_to_3D}
\end{table*}

To predict the displacement vector $\{\Delta \bs{p}_i\}$ for each point, we follow Wang et al.\cite{wang2019dynamic} to extract point features from multiple input point set $P$ through the graph attention modules, which forms a three-stages point feature learning framework. 
At $i$\textsuperscript{th} stage, the deformation pipe takes both the geometric styles $\{\bs{\sigma}_i,\bs{\mu}_i\}$ and semantic feature $\bs{s}_i$ as input, and infers the displacement for each point, according to the geometric information and semantic attribute interpreted from the image feature. 
For convenience, we denote the point features generated in stage $i$ as $\mathcal{Q}^i=\{\bs{q}^{i}_k\}$.

For the geometric styles $\{\bs{\sigma}_i,\bs{\mu}_i\}$, we follow the practice of style transfer\cite{karras2019style} to introduce the adaptive instance normalization, which is used to adapt point features according to the geometric information encoded in the geometric styles. The formulation is given as:
\begin{equation}\small
\hat{\bs{q}}_k^i = \bs{\sigma}_{ik}\cdot {\frac{\bs{q}_k^i-\mu(\bs{q}_k^i)}{\sigma(\bs{q}_k^i)} } + \bs{\mu}_{ik},
\end{equation}
where $\mu(\bs{q}_k^i)$ and $\sigma(\bs{q}_k^i)$ denote the mean and deviations of $\bs{q}_k^i$ estimated by moving average algorithm, respectively. $\bs{\sigma}_{ik}$ and $\bs{\mu}_{ik}$ denote the vector at $k$\textsuperscript{th} row of $\bs{\sigma}_{i}$ and $\bs{\mu}_{i}$, respectively.

After the adaptation of point feature according to the geometric styles, the semantic feature $\bs{s}_i$ is integrated into the point feature $\hat{\bs{q}}_k^i$ through MLP layer and element-wise add, given as: 
\begin{equation}
    \hat{\bs{q}}^{i}_k \leftarrow \hat{\bs{q}}^{i}_k + \phi(\bs{s}_i|\theta_{\bs{s}_i}).
\end{equation}

At the top of the deformation pipe, we use MLP layers to transform point features into 3-dimensional displacement vectors $\{\Delta \bs{p}_k\}$, and finally output the deformed shape as $\{\bs{p}_k+\Delta \bs{p}_k\}$.

\begin{figure}[!t]
  \centering
  \includegraphics[width=0.9\columnwidth]{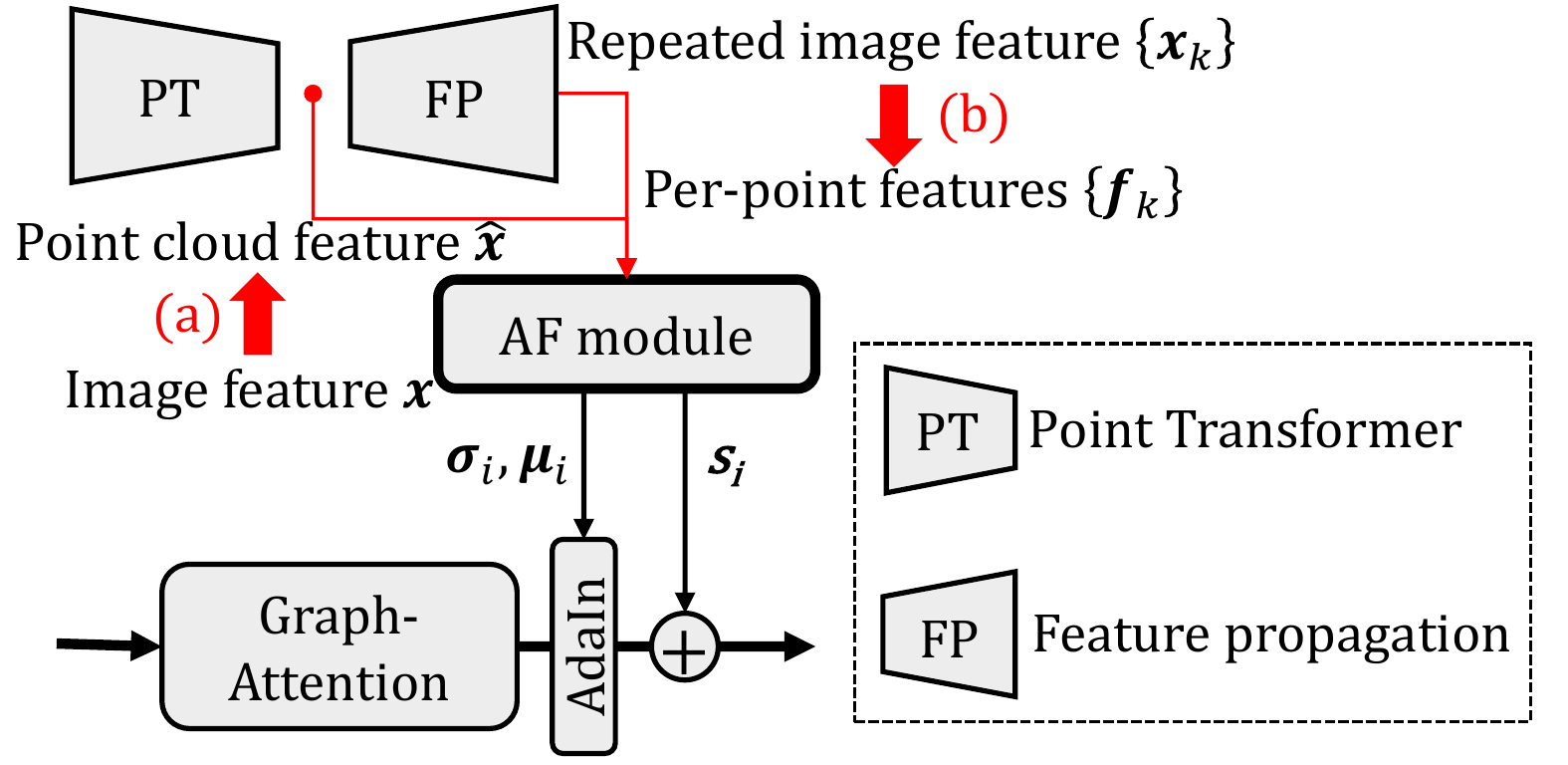}\vspace{-0.2cm}
  \caption{Illustration of extending 3DAttriFlow to 3D shape completion task. This is achieved by (a) replacing the input of attribute flow pipe with the global point cloud feature learned by the PointTransformer, and (b) by replacing the image feature with the per-point feature learned by feature propagation module, respectively. 
  }
  \label{fig:extension}
\end{figure}

\subsection{Extension to Shape Completion}
3DAttriFlow can also be used to predict the missing part of an incomplete shape, which can be achieved by replacing the image encoder with the 3D point cloud encoder (such as PointTransformer\cite{zhao2021point}) in the attribute flow pipe. As a result, the input image feature $\bs{x}$ is replaced by the point cloud feature $\hat{\bs{x}}$. Inspired by PMP-Net\cite{wen2021pmp}, we find that the per-point features of incomplete point cloud can be used as the location prior to guide the move-based completion. Therefore, we replace the repeated image feature $\{\bs{x}_k\}$ with the per-point features $\{\bs{f}_k\}$, which are learned by the feature propagation module specified in PointNet++\cite{qi2017pointnet2}. After that, we concatenate $\{\bs{f}_k\}$ with sphere point cloud $\{\bs{p}_k\}$ as $\{[\bs{f}_k:\bs{p}_k]\}$.
The modification to the attribute flow pipe is illustrated in Figure \ref{fig:extension}. To further improve the completion performance, we follow the coarse-to-fine strategy adopted by most of the completion methods\cite{pan2021variational,xiang2021snowflakenet} to introduce an additional refining module from VRCNet\cite{pan2021variational}, which aims to refine the detailed shape of predicted point clouds.

\subsection{Training loss}
The orthogonality of $\mathcal{U}_i$ is guaranteed by the regularization of orthogonality loss, which is defined as:

\begin{equation}\label{eq:orth_loss}\small
    \mathcal{L}_{\rm Orth} = \sum_{i \in {1,2,3}} {\| \mathcal{U}_i^{\rm T}\mathcal{U}_i \|}-1.
\end{equation}
The deformed shape conditioned by images and incomplete shapes is regularized by the ground truth point cloud through Chamfer distance (CD) defined as:
\begin{equation}\label{eq:cd}\small
\begin{split}
  \mathcal{L}_{\rm CD}(\mathcal{P}^{\rm o},\mathcal{P}^{\rm t})=&{\frac{1}{2N}}\sum_{\bs{p}^{\rm o}\in \mathcal{P}^{\rm {\rm o}}}\min_{\bs{p}^{\rm t}\in \mathcal{P}^{\rm t}} {\|\bs{p}^{\rm o}-\bs{p}^{\rm t} \|}_2 \\
  &+{\frac{1}{2N}} \sum_{\bs{p}^{\rm t}\in \mathcal{P}^{\rm t}}\min_{\bs{p}^{\rm o}\in \mathcal{P}^{\rm o}}{\|\bs{p}^{\rm t}-\bs{p}^{\rm o} \|}_2.
\end{split}
\end{equation}

The total training loss is formulated as
\begin{equation}\label{eq:total_loss}\small
    \mathcal{L} = \mathcal{L}_{\rm CD} + \alpha\mathcal{L}_{\rm Orth},
\end{equation}
where $\alpha$ is a balance factor to determine the weight of $\mathcal{L}_{\rm Orth}$. In this paper, $\alpha$ is set to 100 for all experiments.

\begin{table*}[!t]\small
\setlength{\abovecaptionskip}{-0.cm}
\centering
\caption{Point cloud completion on MVP dataset in terms of per-point L2 Chamfer distance $\times 10^{4}$ (lower is better).}
\resizebox{\textwidth}{!}{
\begin{tabular}{l|c|cccccccccccccccc}
\toprule
Methods &Average  &Plane    &Cabinet   &Car   &Chair   &Lamp    &Sofa  &Table  &Water. &Bed &Bench  &Shelf  &Bus    & Guitar &Motor. &Pistol &Skate.  \\ \midrule
PCN  \cite{yuan2018pcn}   &9.80     &4.22 &8.92    &6.49    &12.46    &19.54    &9.92    &12.45    &8.78    &19.0 &9.0    &13.39  &5.15   &1.87   &6.03   &6.04   &4.70   \\
TopNet  \cite{tchapmi2019topnet}    &10.34   &4.09    &9.71  &7.36   &13.46  &20.53  &11.21  &12.46  &8.50   &18.98  &8.58  &15.15  &5.47   &2.13   &7.19   &7.33   &4.15   \\
MSN \cite{liu2020morphing}  &7.98 &2.59   &8.86   &6.54   &10.22  &12.64  &9.08   &9.69   &7.08   &15.58  &6.38   &11.31 &5.23   &1.37   &4.63   &4.72   &3.06   \\
CRN \cite{wang2020cascaded}  &7.34 &2.45   &8.62   &5.97   &8.95   &11.16  &8.63   &9.30   &6.43   &14.93  &6.11   &10.39 &4.97   &1.67   &4.33   &4.47   &3.39      \\
VRCNet \cite{pan2021variational}  &5.96  &2.17   &7.83   &5.52   &7.31   &8.29   &7.42   &7.07   &5.15   &11.18  &4.76   &7.03   &4.40   &1.15   &3.75   &3.54   &2.31   \\
PMPNet \cite{wen2021pmp}  &6.24  &1.99   &8.84   &6.36   &7.77   &6.18   &8.72   &7.71   &5.19   &11.77  &5.07   &8.34   &5.27   &1.27   &3.95   &3.57   &2.35   \\
SnowflakeNet \cite{xiang2021snowflakenet}  &5.86    &2.04   &7.76   &5.61   &7.07   &7.42   &6.92   &7.13   &5.05   &11.32  &4.87   &7.72   &4.46   &1.16   &3.94   &3.52   &3.64   \\
\midrule
3DAttriFlow(Ours) &\textbf{5.06}    &\textbf{1.59}  &\textbf{7.40}  &\textbf{5.44}  &\textbf{6.05}  &\textbf{5.01}  &\textbf{6.81}  &\textbf{6.14}  &\textbf{4.25}  &\textbf{10.62}  &\textbf{3.73}  &\textbf{6.53}  &\textbf{4.30}  &\textbf{0.95}  &\textbf{3.27}  &\textbf{2.78}  &\textbf{1.78} \\
\bottomrule
\end{tabular}
}
\label{table:completion}
\end{table*}
\section{Experiments}
In this section, we experimentally evaluate the effectiveness of 3DAttriFlow in 2D-to-3D reconstruction task, and analyze its generalization ability through point cloud completion task. The ablation studies will focus on the effectiveness of each part of 3DAttriFlow, and visually analyze the extracted semantic attributes by shape manipulation. 
\subsection{2D-to-3D Reconstruction on ShapeNet dataset}
\label{sec:reconstruction}
\textbf{Dataset briefs and evaluation metric.} We follow the experimental settings of OccNet\cite{scheder2019occupancy} to evaluate our 3DAttriFlow on ShapeNet dataset\cite{chang2015shapenet}. The whole dataset consists of 43,783 mesh object with 13 categories, which will be divided into training, validation and testing following the same strategy of OccNet\cite{scheder2019occupancy}. Since our method focuses on the reconstruction of 3D point cloud from 2D images, we follow AtlasNet \cite{groueix2018papier} to uniformly sample 30k points on the mesh surface of 3D object as the ground truth for training.
Following previous methods\cite{scheder2019occupancy,wang2018pixel2mesh}, we use L1 Chamfer distance described by Eq. (\ref{eq:cd}) as the evaluation metric. In order to compare with other methods of reconstructing 3D voxel or mesh, we follow OccNet\cite{scheder2019occupancy} to sample 2,048 points from their output surface, and then calculate the L1 Chamfer distance with the ground truth.
As for voxel-based methods, we additionally transfer their voxel output into mesh, then apply the point cloud sampling on the mesh surface.

\begin{figure}[!t]
  \centering
  \includegraphics[width=0.95\columnwidth]{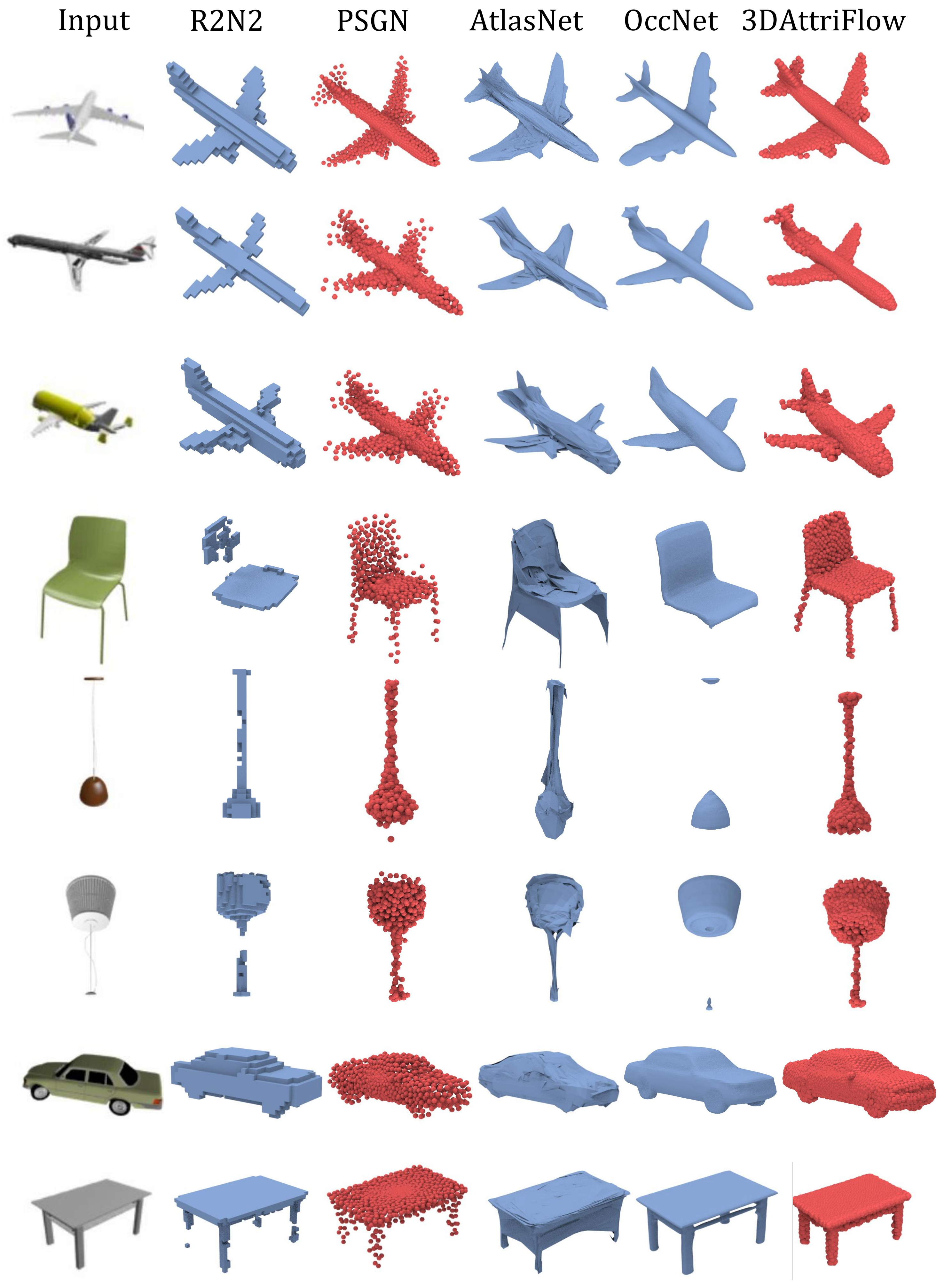}\vspace{-0.2cm}
  \caption{Visual comparison of 2D-to-3D reconstruction results with different methods under ShapeNet dataset.
  }
  \label{fig:2to3_visual_comp}
\end{figure}

\textbf{Quantitative comparison.}
The results of 2D-to-3D reconstruction are shown in Table \ref{table:2D_to_3D}, in which 3DAttriFlow achieves the superior performance over the other compared counterpart methods. Especially, PSGN\cite{fan2017point} and AtlasNet\cite{groueix2018papier} are point cloud based methods, which are most relevant to 3DAttriFlow. However, 3DAttriFlow achieves more than 25\% performance gain over these two methods. As we discussed in Sec. \ref{sec:introduction}, the above-mentioned two methods adopt the typical paradigm for 2D-to-3D reconstruction, where AtlasNet\cite{groueix2018papier} directly decodes the whole shape based on the implicit input of global feature, and PSGN\cite{fan2017point} exploits the feature channels between encoder and decoder for introducing various levels of semantics. None of these practices can learn the explicit semantic features from the image, but only tries to decode shapes from implicit global feature or intermediate layers of encoder. In contrast, 3DAttriFlow can exploit both implicit and explicit semantic attributes learned from the image, which is through the geometric sub-pipe and semantic sub-pipe, respectively. As a result, 3DAttriFlow is able to predict the details of 3D shape based on more definite guidance from the explicit semantic attributes, and achieves better performance than its counterparts. 

\textbf{Qualitative comparison.}
The visual comparison of 2D-to-3D reconstruction is shown in Figure \ref{fig:2to3_visual_comp}. Note that for AtlasNet, we follow its original visualization settings to exhibit the reconstructed mesh instead of point cloud. Compared with the other methods, 3DAttriFlow reconstructs the better details on a wide range of object categories. For example, on the chair category (the $5^{th}$ row of Figure \ref{fig:2to3_visual_comp}), the legs are missing in the prediction of OccNet, while the chair predictions of PSGN and AtlasNet are ambiguous and full of noise. As for the plane category, both PSGN and AtlasNet fail to reconstruct the detailed shape of engines in the $1^{st}$ and the $2^{nd}$ row, while OccNet cannot make the correct prediction of engines steadily (failure in the $2^{nd}$ row).

\begin{figure}[!t]
  \centering
  \includegraphics[width=\columnwidth]{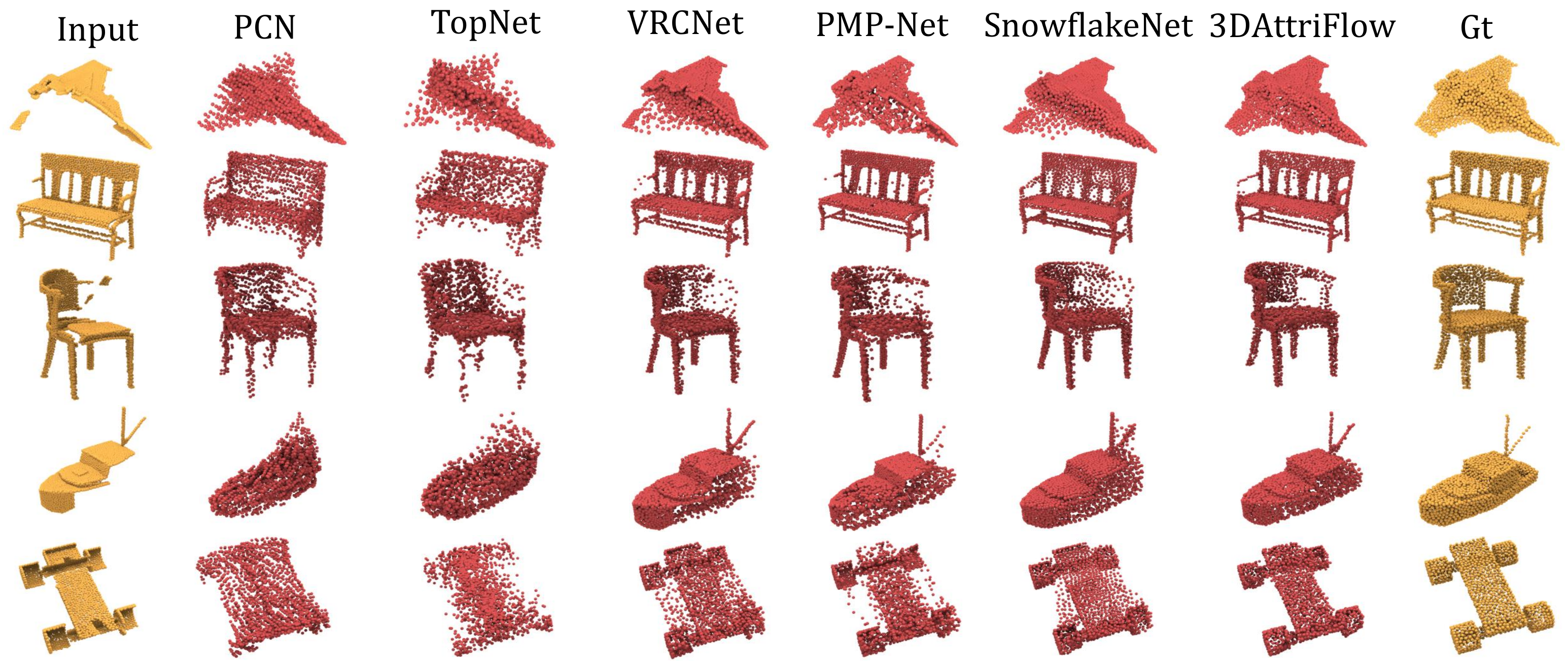}\vspace{-0.2cm}
  \caption{Visual comparison of point cloud completion results with different methods under MVP dataset.
  }
  \label{fig:completion}
\end{figure}

\subsection{3D Completion on MVP dataset}
\textbf{Dataset briefs and evaluation metric.} We follow the experimental settings of VRCNet \cite{pan2021variational} to evaluate our 3DAttriFlow on MVP dataset \cite{pan2021variational}. The dataset consists of 16 categories of incomplete/complete point clouds generated by models selected from ShapeNet, and is then  divided into training set (62,400 shape pairs), and testing set (41,600 shape pairs). Following previous methods \cite{tchapmi2019topnet, pan2021variational, wen2021pmp}, we use L2 Chamfer distance as the evaluation metric.

\textbf{Quantitative comparison.}
The comparison of point cloud completion is shown in Table \ref{table:completion}. Compare with the current state-of-the-art completion method SnowflakeNet\cite{xiang2021snowflakenet}, 3DAttriFlow further improves the performance by 13.7\% in terms of L2-CD. The intuition behind the completion task is the same as the 2D-to-3D reconstruction task, which is to predict a 3D shape based on a given input. In the case of point cloud completion, the input is the incomplete 3D shape. The better performance achieved by 3DAttriFlow can be dedicated to more comprehensive and explicit understanding about the semantic attributes, which is through the semantic sub-pipe in AF module. For example, in order to infer the length of an missing chair legs, a semantic code explicitly controlling such attribute is able to guide the decoder to make a more precise prediction. In contrast, for the compared methods in Table \ref{table:completion}, their decoders have to make the prediction from the implicit features, where the attribute of legs are entangled with the others in the implicit features.

\textbf{Qualitative comparison.}
In Figure \ref{fig:completion}, we qualitatively compare 3DAttriFlow with the other completion methods on MVP dataset, from which we can find that 3DAttriFlow produces more precise and consistent complete shapes than other methods. Take the completion of chairs in the 2\textsuperscript{nd} and the 3\textsuperscript{rd} rows as examples, the predictions of chair-back and the missing chair armrest made by 3DAttriFlow are apparently better than the other methods. As for the skateboard in the 5\textsuperscript{th} row, all of the five compared methods mess the wheels with the board, while 3DAttriFlow can produce a clean and detailed shape of the target skateboard. 

\begin{figure*}[!t]
  \centering
  \includegraphics[width=\textwidth]{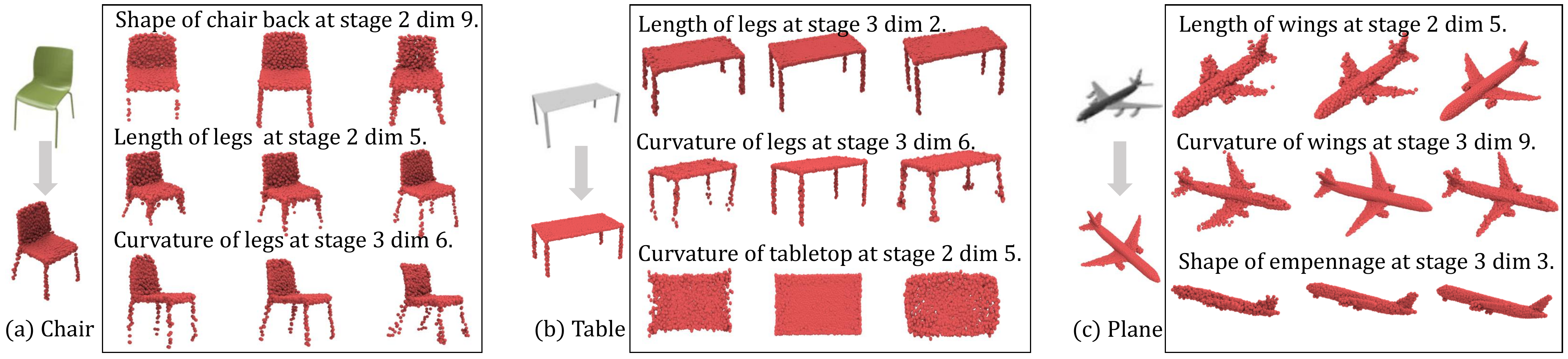}\vspace{-0.2cm}
  \caption{Visualization of 3D shape manipulation through semantic code $\bs{z}$. Each row of the sub-figure shows corresponding shape deformation results caused changing the value of a single dimension of semantic code $\bs{z}$, which proves that semantic code is able to control explicit semantic attributes during the 2D-to-3D reconstruction process. Since we cannot control which semantic attribute is encoded by network, we manually go through the dimensions of semantic code $\bs{z}$ and reveal the learned semantic attributes.
  }
  \label{fig:manipulation}
\end{figure*}

\subsection{Ablation Studies}
In this subsection, all quantitative analysis results are typically conducted under four categories (i.e. plane, car, chair and table). By default, all the experimental settings are kept the same as in Sec. \ref{sec:reconstruction}, except for modified part described in each ablation experiment below.

\textbf{Analysis of each sub-pipe in AF module.}
We analyze the effectiveness of each sub-pipe of 3DAttriFlow by removing/replacing modules from the original network structure (denoted as \emph{Full}). 
Specifically, we develop four different variations for comparison: (1) \emph{w/o semantic sub-pipe} is the variation removing semantic sub-pipe from the AF module; (2) \emph{w/o geometric sub-pipe} is the variation removing geometric sub-pipe from the AF module; (3) \emph{semantic MLPs} is the variation replacing semantic sub-pipe with simple MLP layer, where the output is directly added to the features in deformation pipe; (4) \emph{geometric MLPs} is the variation replacing geometric sub-pipe with simple MLPs, where the output is added to the features in deformation pipe.
The results are shown in Table \ref{table:part_analysis}, from which we can find that our Full model achieves the best results over all four variations. Such result proves the effectiveness of each part to 3DAttriFlow. 

Moreover, we additionally address two conclusions. 
First, by comparing w/o geometric sub-pipe and w/o semantic sub-pipe to the Full model, we can find that semantic sub-pipe has a relatively less impact on the performance of 2D-to-3D reconstruction than geometric sub-pipe. 
The reason is that, although semantic sub-pipe can explicitly disentangle and extract the semantic attribute from 2D images, there always exist certain semantic attributes that cannot be explicitly captured or disentangled. Therefore, an implicit representation is still necessary for encoding such implicit semantic attributes in images. Second, by comparing geometric-MLP and semantic-MLP to the only-MLPs, we can find that both geometric sub-pipe and semantic sub-pipe are more effective than simple MLPs, which proves the effectiveness of the network designation of the two sub-pipes.

\begin{table}[!h]\small
\setlength{\abovecaptionskip}{-0.cm}
\centering
\caption{The effect of each sub-pipe to 3DAttriFlow in terms of L1-CD$\times 10^2$.}
\resizebox{\columnwidth}{!}{
\begin{tabular}{lccccc}
\toprule
Steps. &avg.    &plane   &car  &chair  &table  \\ \midrule
w/o semantic sub-pipe  &3.16  &2.58  &2.80  &3.89 &3.35\\
w/o geometric sub-pipe  &3.41 &2.66  &3.07  &4.23  &3.68 \\
semantic-MLP  &3.12  &2.53  &2.85  &3.81 &3.30 \\
geometric-MLP  &3.08  &\textbf{2.47}  &2.73  &3.80 &3.30 \\
only-MLP &3.21  &2.67  &2.82  &3.91 &3.45\\
Full &\textbf{3.03}  &2.49  &\textbf{2.69}  &\textbf{3.73} &\textbf{3.23} \\
\bottomrule
\end{tabular}
}
\label{table:part_analysis}
\end{table}

\textbf{Visualization of semantic attributes controlled by semantic code $\bs{z}$.} The semantic code $\bs{z}$ is expected to encode explicit semantic attribute into the activation of a single dimension, which aims to provide definite guidance for the reconstruction of 3D shape. 
In order to visually analyze the encoded semantic attributes captured by $\bs{z}$, we go through the dimensions of $\bs{z}$ and observe the shape deformations caused by interpolating single dimensions of $\bs{z}$, as shown in Figure \ref{fig:manipulation}. 
Specifically, we illustrate our observations of 3 attributes for each of the 3 categories, which proves that semantic code $\bs{z}$ successfully captures the explicit semantic attribute, and effectively reveals the reconstruction of the corresponding part of 3D shape.
For example, as for the reconstruction of chair (Figure \ref{fig:manipulation}(a)), the code $\bs{z}$ learns two specific semantic attributes of legs, which are the bending (encoded by 6\textsuperscript{th} dimension at stage 2) and the length (encoded by the 5\textsuperscript{th} dimension at stage 3), respectively. From the visualization results of Figure \ref{fig:manipulation}(c), we can find that changing the value of activation will result in obvious deformation of corresponding semantic attributes. Moreover, by observing the extracted semantic attributes across three categories, we can find that semantic code $\bs{z}$ is able to generalize its learned attributes into multiple categories, as the same attributes of bending and length can also be found in the table and plane categories.

\begin{table}[!h]\small
\setlength{\abovecaptionskip}{-0.cm}
\centering
\caption{The effect of code dimension in terms of L1-CD$\times 10^2$ (baseline marked by ``*'').}
\begin{tabular}{lccccc}
\toprule
Dims. &avg.    &plane   &car  &chair  &table  \\ \midrule
4&3.14  &2.60  &2.81  &3.84 &3.30\\
8  &3.11  &2.51  &2.73  &3.83 &3.35\\
18*  &\textbf{3.03}  &\textbf{2.49}  &\textbf{2.69}  &\textbf{3.73} &\textbf{3.23}\\
32  &3.20   &2.58  &2.75  &3.85  &3.34 \\
\bottomrule
\end{tabular}
\label{table:dim_analysis}
\end{table}

\begin{figure}[!h]
  \centering
  \includegraphics[width=0.95\linewidth]{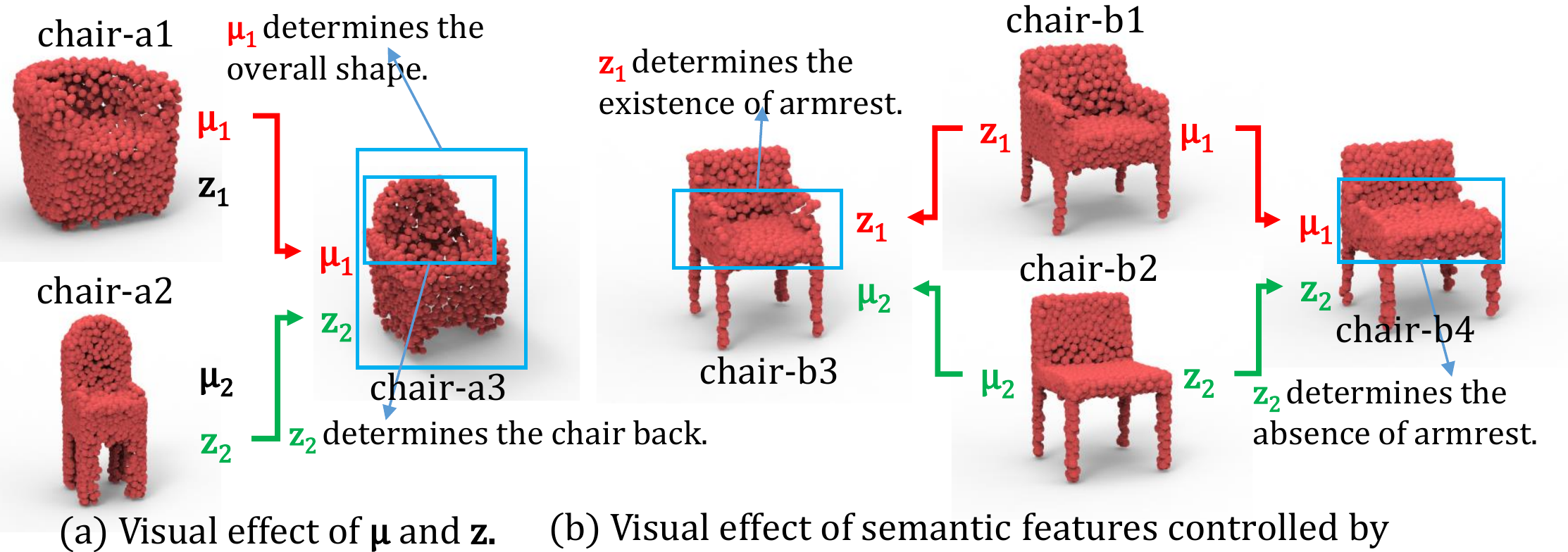}\vspace{-0.2cm}
  \caption{Effect of replacing semantic/geometric code $z$/$\mu$. In (a), we replace the semantic code $z_1$ of chair-a1 with the code $z_2$ from chair-a2. The output chair-a3 shows that geometric code $\mu_1$ controls the overall shape, and the semantic code $z_2$ controls the shape of chair back. In (b), we further compare the semantic attributes controlled by different semantic code $z$. Chair-b3 in (b) inherits the armrest from the chair-b1 through semantic code $z_1$, while chair-b4 drops the armrest according to the semantic code $z_2$ from chair-b2.
  }
  \label{fig:codez}
\end{figure}

\textbf{Analysis of semantic code $\bs{z}$.}
Since each dimension of semantic code $\bs{z}$ can potentially encode a certain semantic attribute, in this part, we discuss the capability of semantic code $\bs{z}$ for encoding semantic attributes in terms of code dimensions. We report the results under code dimension of 4, 8 and 32 following the power of 2, and compare it with our default setting 18 in Table \ref{table:dim_analysis}. From the results we can find that with 18-dimensional semantic code 3DAttriFlow achieves the best performance, while the other settings cause a relatively small performance drop. The reason is that for small dimensions, the semantic code can only encode the limited semantic attributes, which is insufficient for predicting a detailed 3D shape. On the other hand, large dimensions may have the problems for learning orthogonal bases to represent the semantic attributes. Moreover, we visualize the effect for swapping code $z$ and $\mu$ from different objects in Figure \ref{fig:codez}, from which we can observe that several geometric/semantic attributes are clearly controlled by code $\mu$ and $z$, respectively.

\vspace{-0.2cm}


\section{Conclusions and Limitations}
In this paper, we propose 3DAttriFlow to reconstruct 3D shapes from 2D images. Compared with the previous methods, which merely learn to reconstruct the 3D shape based on the implicit features, 3DAttriFlow takes the advantage of a novel attribute flow pipe to explicitly extract semantic attributes from the implicit feature, which makes the 3D shape prediction more accurate based on the extracted semantic attributes. To overcome the problem of generating discrete point cloud data, the deformation pipe is proposed to combine with the attribute pipe, which provides location priors for the extracted semantic attributes. Comprehensive experiments on ShapeNet dataset for 2D-to-3D reconstruction and MVP dataset for point cloud completion have proved the effectiveness of 3DAttriFlow, and the visualization of shape manipulation also demonstrates the ability of 3DAttriFlow to extract and control the explicit semantic attributes of 3D shapes.

The limitations and possible future work of 3DAttriFlow can be addressed as follows. Although the semantic code $\bs{z}$ is able to learn the explicit semantic attributes and encode them into certain dimensions, it cannot always learn a meaningful or disentangled semantic attributes for every dimension. In experiments, we observe that some dimensions may have effect to several attributes, while others may have little effect on the output shape. In our opinion, this can be dedicated to the information loss/compression during the extraction process of global image feature, which may cause the semantic attribute missing or deeply entangled with each other. Therefore, the feature channels connecting multiple layers of encoder to the attribute flow pipe is still necessary, in order to fully utilize the ability of semantic attribute extraction of 3DAttriFlow. 
{\small
\bibliographystyle{ieee_fullname}
\bibliography{egbib}
}

\end{document}